\documentclass[sigconf]{acmart}
\AtBeginDocument{%
  }

\usepackage{amsmath, amsthm}
\newtheorem{observation}{Observation}
\newcommand{\insight}[1]{\noindent\textbf{Insight:} \textit{#1}}
\usepackage{multirow}
\usepackage{subcaption}
\usepackage{graphicx}            

\begin{document}

\title{Fairness Begins with State: Purifying Latent Preferences for Hierarchical Reinforcement Learning in Interactive Recommendation}

\author{Yun Lu}
\email{luyun@cigit.ac.cn}
\affiliation{
  \institution{Chongqing Institute of Green and Intelligent Technology, Chinese Academy of Sciences}
  \institution{Chongqing School, University of Chinese Academy of Sciences}
  \city{Chongqing}
  \country{China}
}

\author{Xiaoyu Shi}
\email{xiaoyushi@cigit.ac.cn}
\affiliation{
  \institution{Chongqing Institute of Green and Intelligent Technology, Chinese Academy of Sciences}
  \institution{Chongqing School, University of Chinese Academy of Sciences}
  \city{Chongqing}
  \country{China}
}

\author{Hong Xie}
\authornotemark[3]
\email{xiehong2018@foxmail.com}
\affiliation{%
  \institution{The First Affiliated Hospital, University of Science and Technology of China}
  \city{Hefei}
  \country{China}
}

\author{Xiangyu Zhao}
\email{xy.zhao@cityu.edu.hk}
\affiliation{%
  \institution{City University of Hong Kong}
  \city{Hong Kong}
  \country{Hong Kong}
}

\author{Mingsheng Shang}
\email{mingshengshang@cigit.ac.cn}
\affiliation{%
  \institution{Chongqing Institute of Green and Intelligent Technology, Chinese Academy of Sciences}
  \institution{Chongqing School, University of Chinese Academy of Sciences}
  \city{Chongqing}
  \country{China}
}

\begin{abstract} 
Interactive recommender systems (IRS) are increasingly optimized with Reinforcement Learning (RL) to capture the sequential nature of user-system dynamics. However, existing fairness-aware methods often suffer from a fundamental oversight: they assume the observed user state is a faithful representation of true preferences. In reality, implicit feedback is contaminated by popularity-driven noise and exposure bias, creating a distorted state that misleads the RL agent. We argue that the persistent conflict between accuracy and fairness is not merely a reward-shaping issue, but a state estimation failure. In this work, we propose \textbf{DSRM-HRL}, a framework that reformulates fairness-aware recommendation as a latent state purification problem followed by decoupled hierarchical decision-making. We introduce a Denoising State Representation Module (DSRM) based on diffusion models to recover the low-entropy latent preference manifold from high-entropy, noisy interaction histories. Built upon this purified state, a Hierarchical Reinforcement Learning (HRL) agent is employed to decouple conflicting objectives: a high-level policy regulates long-term fairness trajectories, while a low-level policy optimizes short-term engagement under these dynamic constraints. Extensive experiments on high-fidelity simulators (KuaiRec, KuaiRand) demonstrate that DSRM-HRL effectively breaks the "rich-get-richer" feedback loop, achieving a superior Pareto frontier between recommendation utility and exposure equity.
\end{abstract}


\maketitle
\section{Introduction}

Interactive recommender systems (IRS) have become the bedrock of modern digital ecosystems, powering platforms from short-video streaming to large-scale e-commerce~\cite{b1,b2}. To navigate the complex, sequential dynamics of user-system interactions, Reinforcement Learning (RL) has emerged as the dominant paradigm~\cite{b3}. By optimizing for long-term cumulative rewards, RL-based agents can theoretically adapt to evolving user preferences and maximize platform utility over extended horizons ~\cite{b4}. As these systems exert increasing influence over information dissemination, the issue of recommendation fairness has become a critical concern~\cite{b5, b6}. While RL-based agents effectively maximize long-term utility, they often inadvertently exacerbate the "rich-get-richer" phenomenon, leading to severe item-side exposure unfairness~\cite{b6}. In this work, we specifically focus on item-side exposure fairness, aiming to rectify the disproportionate visibility of popular items while ensuring that long-tail items with high latent utility receive equitable opportunities for exposure.

Most existing fairness-aware RL methods attempt to mitigate these biases at the decision level—either by augmenting rewards with fairness penalties or by applying constrained optimization to the policy output~\cite{b7, b8}. However, we argue that these approaches suffer from a fundamental oversight: they assume the observed user state is a faithful representation of true preferences. In reality, implicit feedback is heavily contaminated by epistemic uncertainty~\cite{b9}, where signals are warped by popularity-driven noise rather than genuine interest. As illustrated in Figure 1 (Left), user signals are "contaminated" by interface bias and social conformity, creating a significant reliability gap. When fairness interventions are applied to such a distorted state, the agent faces an artificial conflict between accuracy and equity (\textbf{Figure 1, Middle}). Consequently, the conflict is not merely a reward design problem; it is fundamentally a state estimation failure.
\begin{figure*}[t]
\centering
\includegraphics[width=\textwidth]{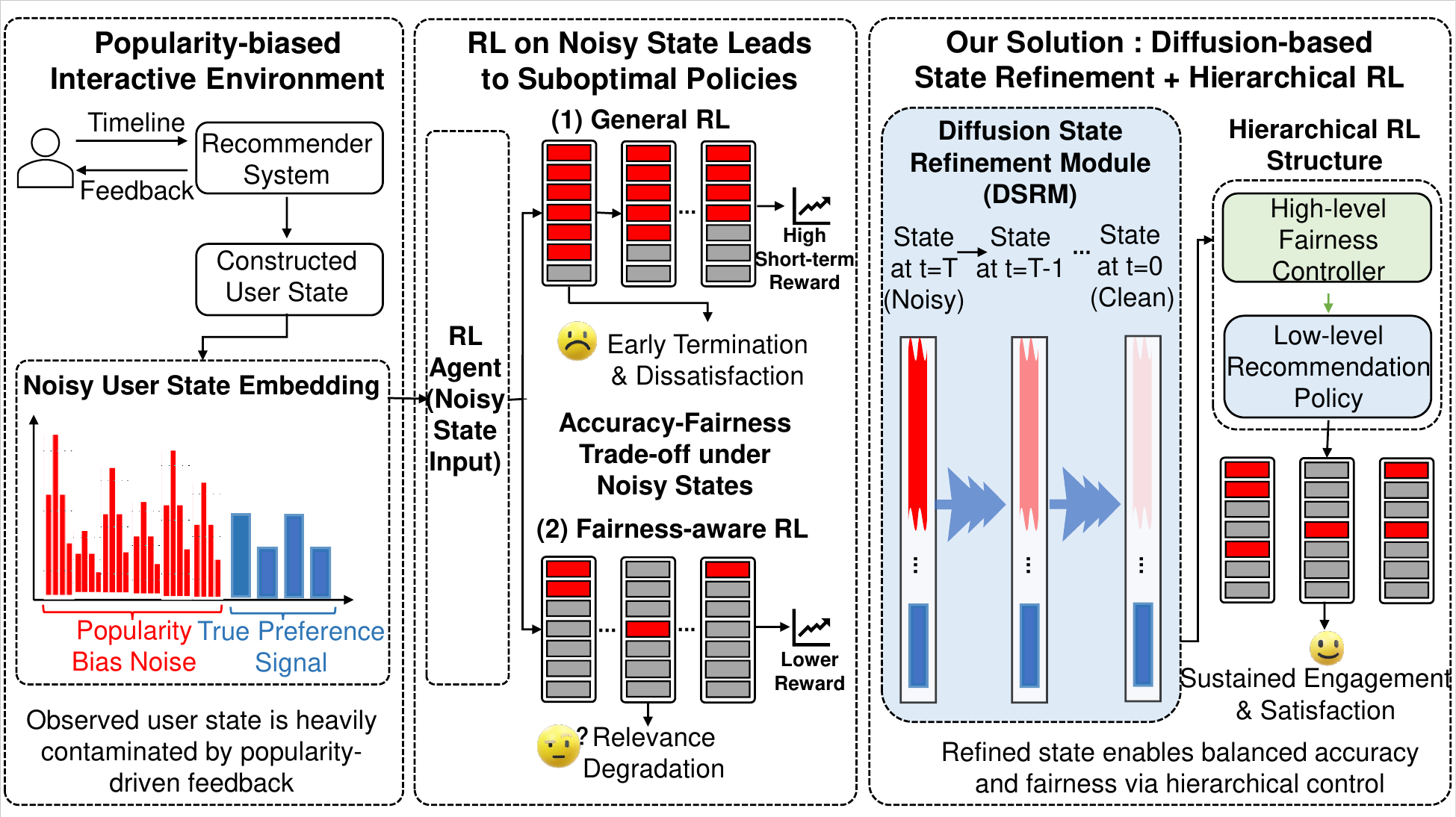} 
\caption{The motivation of DSRM-HRL. \textbf{(Left)} In interactive environments, the observed user state is heavily contaminated by popularity bias (red noise), obscuring true user preferences. \textbf{(Middle)} Existing RL methods typically fail under such noise: general RL agents succumb to the "rich-get-richer" loop (high AD), while naive fairness-aware agents sacrifice accuracy for diversity. \textbf{(Right)} Our proposed DSRM-HRL first purifies the state via diffusion-based denoising and then employs a hierarchical policy to dynamically balance long-term fairness and short-term utility.}
\label{fig:motivation}
\end{figure*}

To resolve this, we reformulate fairness-aware interactive recommendation as a two-stage task: latent state purification followed by decoupled hierarchical decision-making. However, implementing this logic in dynamic environments involves several non-trivial challenges:
\begin{itemize}
    \item \textbf{C1: Complexity of Non-Linear Bias Reconstruction.} Popularity bias is not Gaussian "white noise" but a systematic, non-linear warping of the preference manifold. Traditional denoising autoencoders or linear filters often fail to disentangle genuine niche interests from system-induced popularity fluctuations, leading to "information collapse."

    \item \textbf{C2: Signal Retention vs. Noise Elimination.} Aggressive denoising can inadvertently smooth out the subtle behavioral nuances necessary for personalized recommendation. Recovering a low-entropy latent state without losing the fine-grained "signal" of user intent remains a significant technical bottleneck.

    \item \textbf{C3: Temporal Conflict in Multi-Objective Optimization.} Item-side fairness is a long-term trajectory goal, while recommendation accuracy is an immediate, short-term objective. Training a single-level RL agent to handle both on a corrupted state leads to gradient interference and training instability.
\end{itemize}

To address these challenges, we propose DSRM-HRL (\textbf{Figure 1, Right}), a framework that integrates state purification with Hierarchical Reinforcement Learning (HRL). To tackle C1 and C2, we introduce the Denoising State Representation Module (DSRM). Leveraging the generative power of Diffusion Models, DSRM treats the noisy interaction history as a corrupted signal and reconstructs the underlying latent preference manifold through an iterative reverse process. This ensures high-fidelity state purification that preserves true user intent. To address C3, we employ a Hierarchical RL architecture that achieves temporal decoupling: a high-level policy regulates macro-level fairness across long horizons, while a low-level policy optimizes immediate user engagement conditioned on these fairness constraints. We conduct extensive experiments using KuaiSim $[b17]$, a high-fidelity simulator built on real-world datasets KuaiRec and KuaiRand-Pure. The results demonstrate that DSRM-HRL consistently achieves a superior Pareto frontier compared to both general RL baselines and state-of-the-art fairness-aware methods.

In summary, our main contributions are threefold: 
\begin{itemize} 
\item We identify and formalize a fundamental modeling flaw in fairness-aware recommendation: the assumption of an unbiased user state. We argue that effective fairness intervention must begin with state estimation, not merely reward shaping.
\item We propose a novel architecture that combines diffusion-based state denoising with hierarchical policy decoupling. DSRM reconstructs the decision-relevant preference manifold, while the HRL structure resolves the objective conflict across different temporal scales.
\item We demonstrate that DSRM-HRL significantly improves long-tail item exposure and cumulative user rewards, validating that state purification is a necessary prerequisite for robust fairness-aware decision-making.
\end{itemize}

\section{Related Work}
Our work is closely related to three lines of research: (1) fairness-aware reinforcement learning for recommendation, (2) denoising techniques in recommender systems, and (3) diffusion models for recommendation. We discuss their limitations and position our work accordingly.

\textbf{Fairness-aware IRS.} Recent studies have incorporated fairness objectives into reinforcement learning-based recommender systems by modifying reward functions or introducing additional constraints. Representative methods include multi-objective RL frameworks~\cite{b9} that balance user utility and item exposure, entropy-regularized policies that encourage exploration~\cite{b12}, and diversity- or novelty-aware reward shaping strategies~\cite{b10,b11}. While these approaches differ in implementation, these methods rely on a hidden assumption: the observed state space is a sufficient statistic for true user preferences. We argue that this assumption fails in interactive systems where implicit feedback is systematically warped by exposure bias~\cite{b4}. Unlike these methods that rectify decisions, our work targets the root cause of unfairness: the corrupted input signal itself. By shifting the focus from reward shaping to state purification, we provide a more stable foundation for long-term fairness.

\textbf{Denoising in Recommender Systems.} Existing denoising methods primarily focus on data-level or model-level interventions. Data-level approaches attempt to identify and downweight noisy interactions through resampling or reweighting schemes~\cite{b5,b14}, while model-level approaches employ autoencoders or robust objectives to improve representation stability~\cite{b13}. However, these are largely designed for offline training of static models. In the context of RL, state noise is dynamic and self-reinforcing: a biased recommendation leads to a biased state, which in turn breeds further bias. Our DSRM module differs by performing real-time latent state inference, effectively filtering out popularity-driven fluctuations before they propagate through the decision-making loop.

\textbf{Diffusion Models for Recommendation} Diffusion models have recently been applied to recommendation tasks, mainly for generating user–item interactions or directly sampling recommendation lists~\cite{b15,b16}. They are predominantly used as generative samplers. We pivot this role toward representation purification. Instead of generating new items, our framework uses the reverse diffusion process to recover the low-entropy preference manifold from high-entropy observation noise. This transformation from "generation" to "denoising inference" is key to enabling robust, fair interactions.

\section{Empirical Motivation: The Fallacy of Trustworthy States}
To substantiate our core philosophy that \textbf{fairness begins with state purification}, we empirically investigate whether item-side exposure bias constitutes a fundamental obstacle to interactive recommendation. Our analysis reveals three critical insights: (1) the existence of a popularity-driven \textit{spurious feedback loop}, (2) the \textit{input-side bottleneck} in RL optimization, and (3) the \textit{manifold collapse} within the latent state space.

\subsection{The Spurious Feedback Loop: Why Rewards are Misleading}
\begin{figure}[!t]
    \centering
    \begin{subfigure}{0.48\linewidth}
        \includegraphics[width=\textwidth]{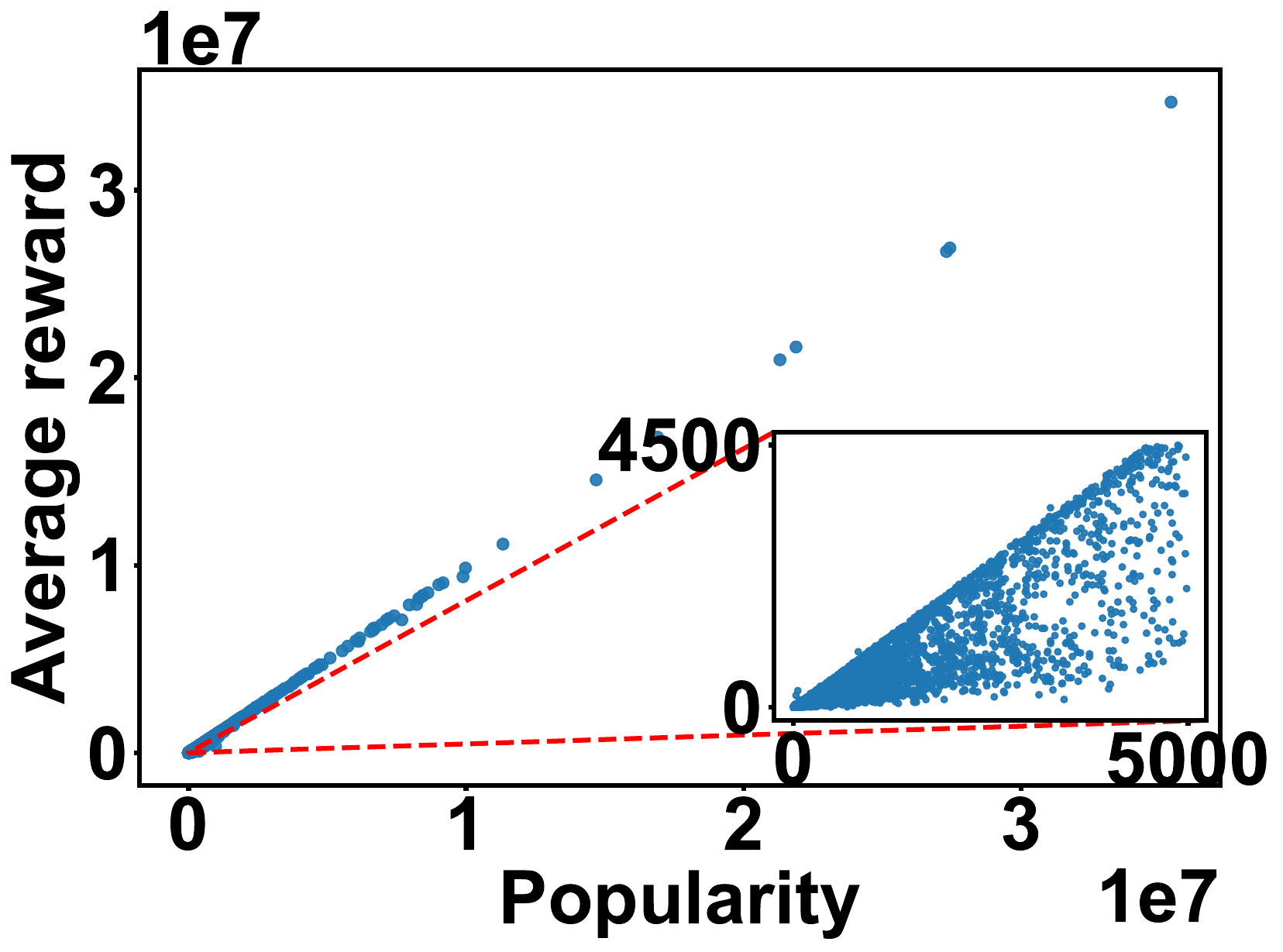}
        \caption{KuaiRec}
    \end{subfigure}
    \begin{subfigure}{0.48\linewidth}
        \includegraphics[width=\textwidth]{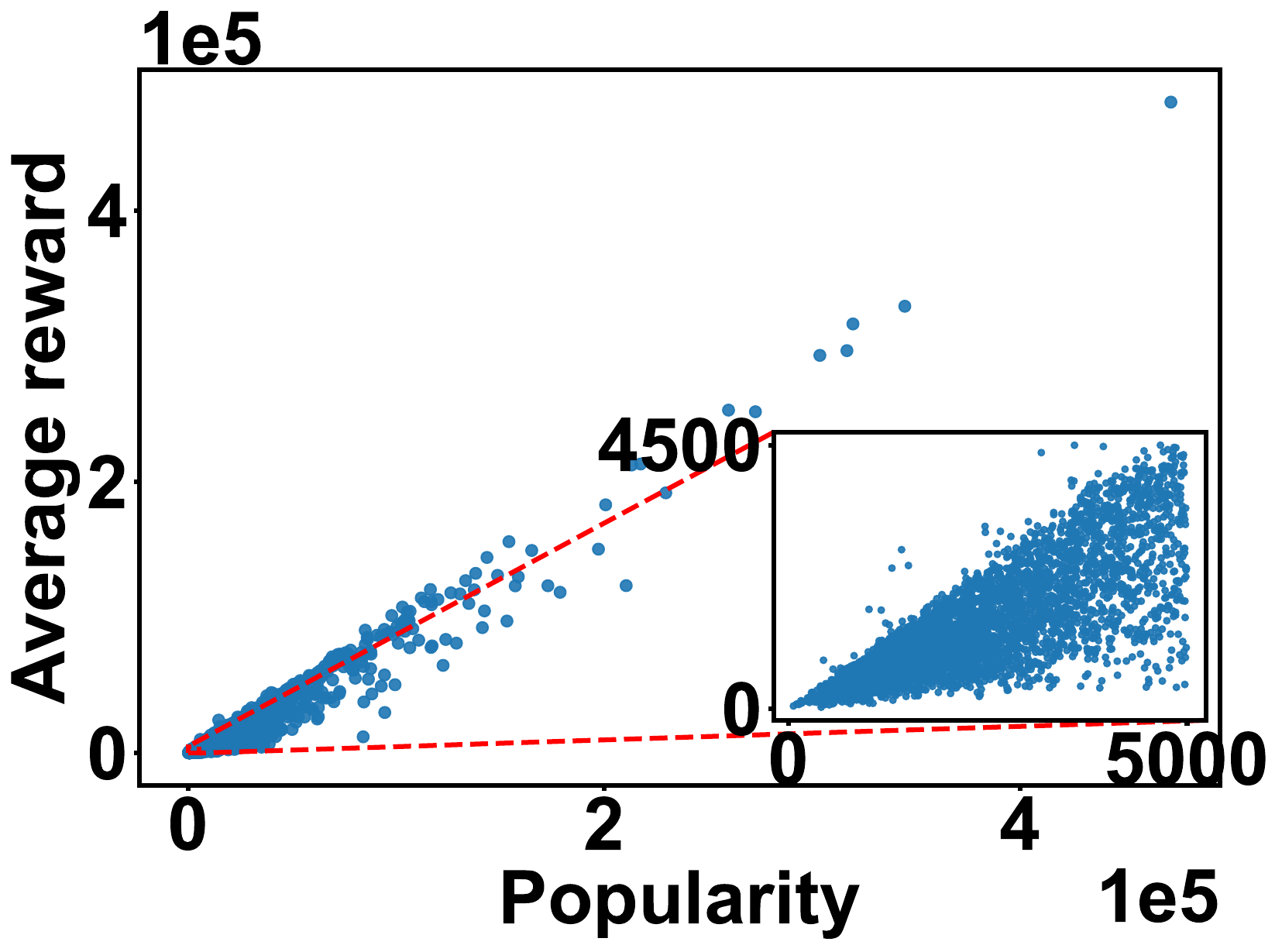}
        \caption{KuaiRand}
    \end{subfigure}
    \caption{The Spurious Reward Trap. Observed rewards are heavily dominated by exposure frequency rather than intrinsic relevance, creating a biased input for policy learning.}
    \label{fig:popularity_reward}
\end{figure}

We first examine whether observed user feedback is dominated by intrinsic preferences or merely a byproduct of exposure frequency. Using \textit{show\_user\_num} as a proxy for popularity $freq(i)$ and \textit{play\_user\_num} for the reward signal, we evaluate KuaiRec and KuaiRand-Pure.
\begin{observation}
As illustrated in Figure~\ref{fig:popularity_reward}, there exists a \textbf{striking linear correlation ($R^2 > 0.85$) between item popularity and average reward}. Highly exposed items systematically harvest higher feedback scores, regardless of their actual utility.
\end{observation}
\insight{This reveals a Spurious Feedback Loop: the system is not learning what users genuinely like, but what it has previously forced them to see. This "popularity trap" misleads the RL agent into recommending "safe" popular items to maximize rewards, reinforcing exposure imbalance. This confirms that implicit feedback is a \textbf{biased proxy} for user intent, validating the \textbf{Epistemic Uncertainty} identified in \textbf{Challenge C1}.}

\subsection{The Input-Side Bottleneck: Denoising vs. Policy Optimization}
\begin{figure}[!t]
    \centering
    \begin{subfigure}{0.48\linewidth}
        \includegraphics[width=\textwidth]{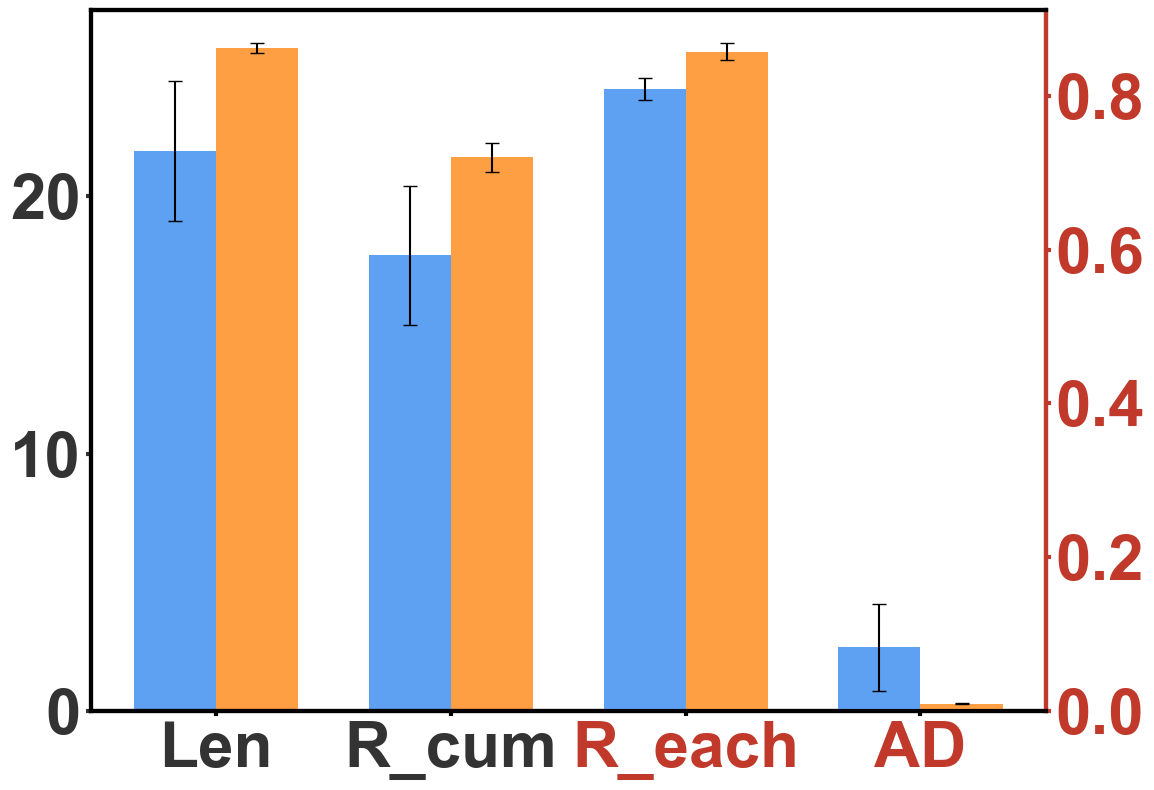}
        \caption{KuaiRec}
    \end{subfigure}
    \begin{subfigure}{0.48\linewidth}
        \includegraphics[width=\textwidth]{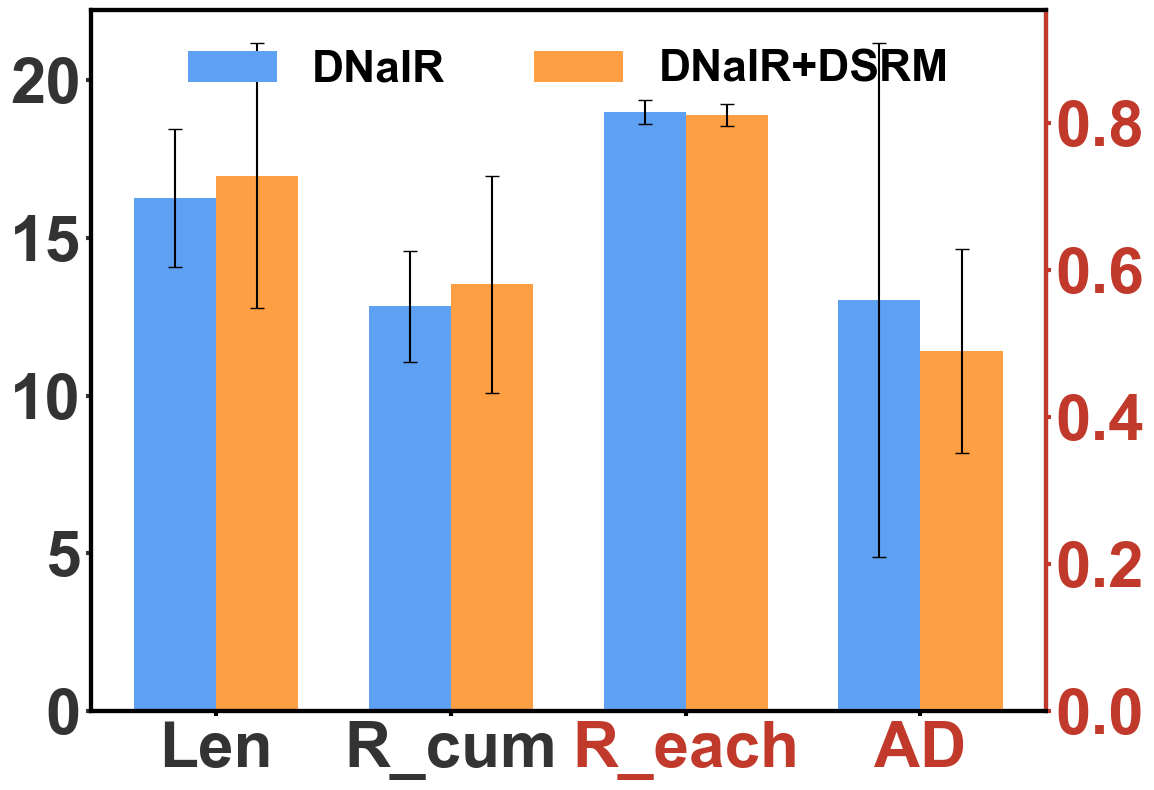}
        \caption{KuaiRand}
    \end{subfigure}
    \caption{State Purification Gain. Denoising the input alone expands the Pareto frontier of accuracy and fairness without complex reward shaping.}
    \label{fig:state_rl}
\end{figure}

To verify if this bias degrades decision-making, we investigate whether the accuracy--fairness conflict originates from the policy or the state. We compare a standard DNaIR policy trained on \textit{Raw States} vs. \textit{Denoised States} purified by our DSRM module.
\begin{observation}
Figure~\ref{fig:state_rl} reveals a \textbf{paradigm-shifting result}: simply purifying the state representation, \textit{without changing the policy or reward function}, achieves a \textbf{simultaneous improvement in accuracy and equity}. On KuaiRec, DSRM yields an 88\% improvement in Absolute Difference (AD) and an 18.4\% gain in interaction length.
\end{observation}

\insight{This proves the accuracy-fairness trade-off is often an artifact of state corruption. When popularity noise is removed, the agent can finally discern the latent utility of long-tail items. This demonstrates that state purification is a necessary prerequisite for resolving \textbf{Challenge C3}, providing a clear signal for the HRL agent to decouple engagement from fairness goals.}

\subsection{Latent Manifold Collapse and Disentanglement}
We analyze the latent geometry using t-SNE, defining \textbf{Popularity labels} (historical exposure) and \textbf{Interest labels} (item category).

\begin{figure}[!t]
    \centering
    \begin{subfigure}{0.96\linewidth}
        \includegraphics[width=\textwidth]{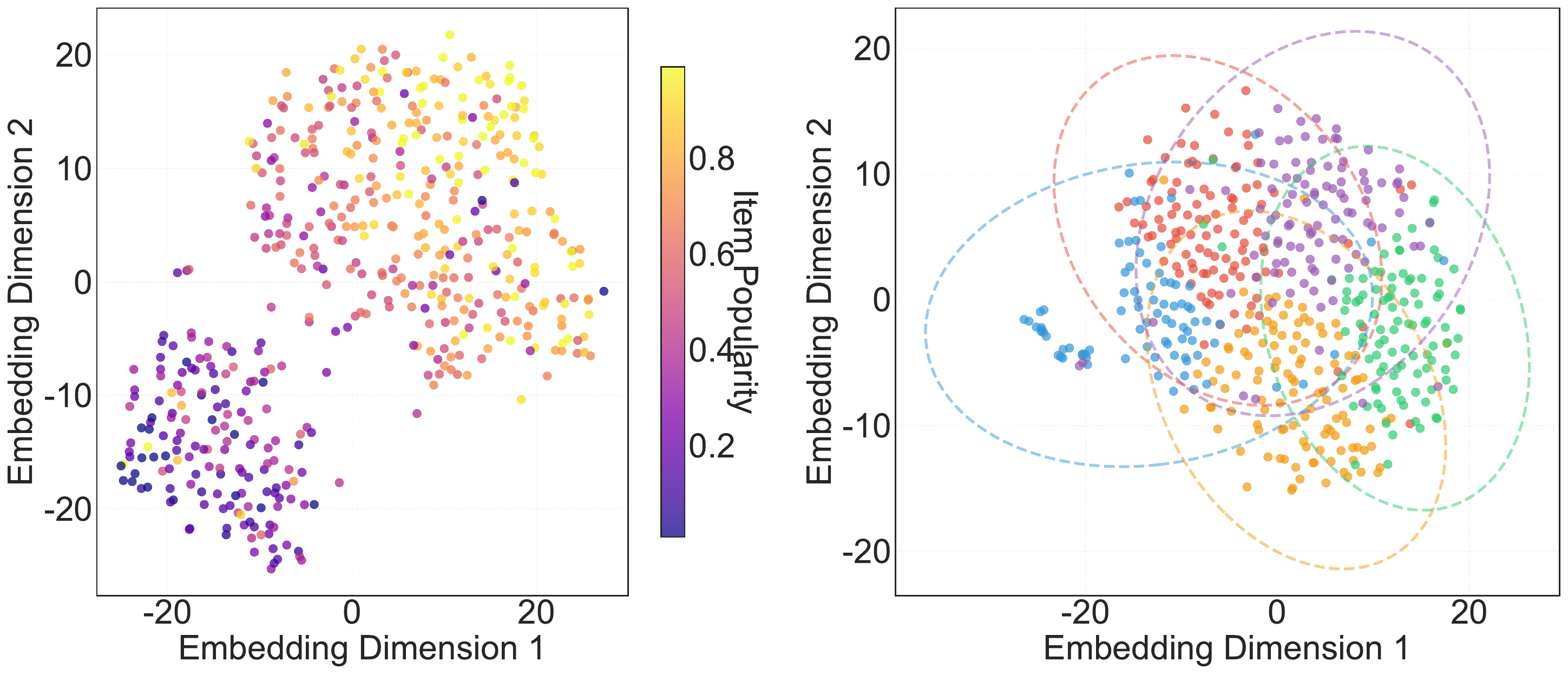}
        \caption{KuaiRec}
    \end{subfigure}
    \begin{subfigure}{0.96\linewidth}
        \includegraphics[width=\textwidth]{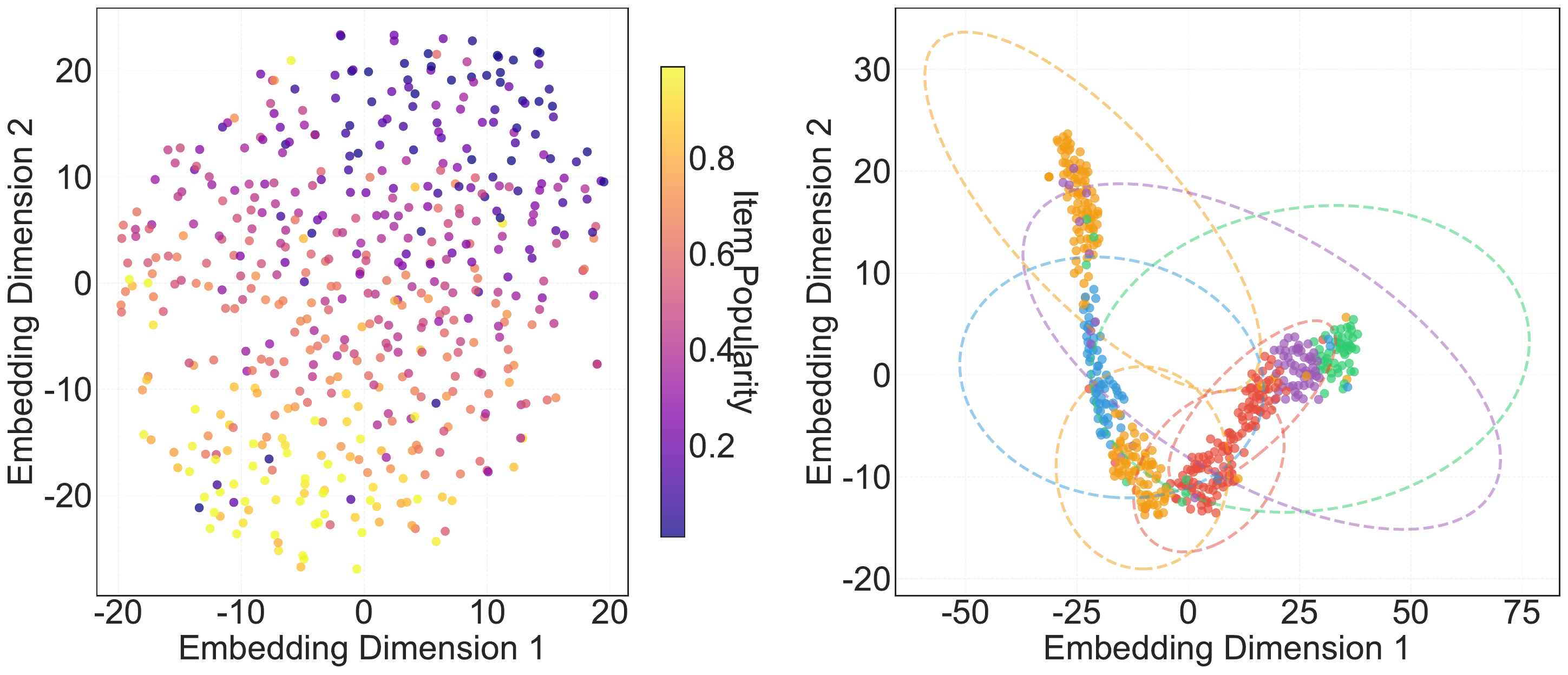}
        \caption{KuaiRand}
    \end{subfigure}
    \caption{Visualization of State Purification. DSRM transforms the popularity-dominated collapsed manifold into a disentangled, semantic preference space.}
    \label{fig:state_vis}
\end{figure}

\begin{observation}
In Figure~\ref{fig:state_vis}, the \textit{Raw State} space exhibits a \textbf{Manifold Collapse}: user embeddings are clustered by popularity levels, while interest categories are heavily \textbf{entangled and indistinguishable}.
\end{observation}

\insight{This provides visual evidence for \textbf{Challenge C2}: popularity noise smothers the semantic preference signal. In contrast, our diffusion-based \textbf{DSRM recovers a Disentangled Preference Manifold}}, where states form semantically consistent clusters. This transition from noise-driven scattering to interest-driven clustering confirms that DSRM's non-linear reconstruction is essential for effective hierarchical control.

Guided by these empirical findings, we next introduce DSRM-HRL, which explicitly models these purified manifolds for robust hierarchical control.

\section{Methodology}
\begin{figure*}[!t]
\centering
\includegraphics[width=\textwidth]{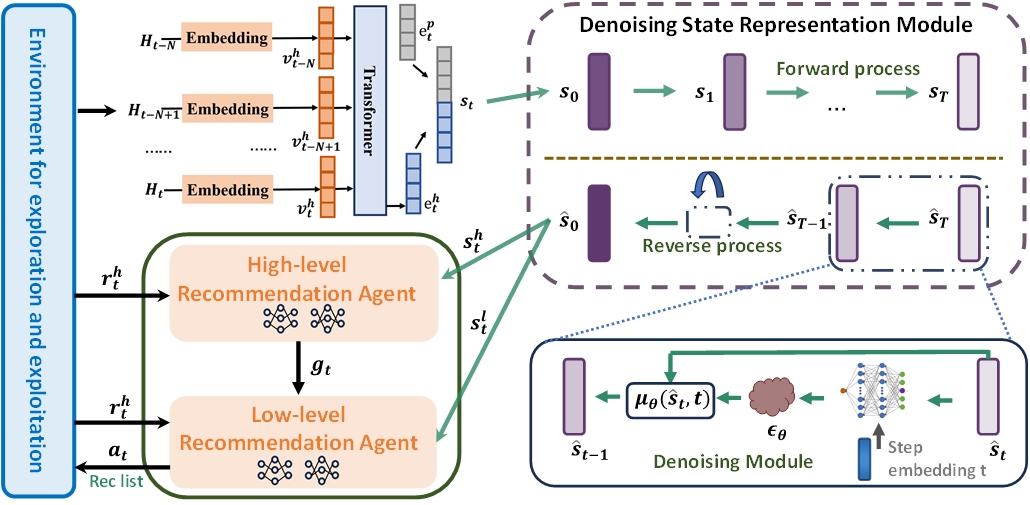}
\caption{Overall framework.}
\label{fig:framework}
\end{figure*}

We formulate fairness-aware interactive recommendation as a problem of 
\textbf{latent state inference followed by hierarchical constrained control}. 
The overall architecture of \textbf{DSRM-HRL} is illustrated in Figure~\ref{fig:framework}.

\subsection{Problem Formulation}

We model the interactive recommendation process as a Markov Decision Process (MDP) 
$(\mathcal{S}, \mathcal{A}, \mathcal{P}, \mathcal{R}, \gamma)$. 
At each time step $t$, the system observes a history-based user state 
$\tilde{s}_t \in \mathcal{S}$, selects an action $a_t \in \mathcal{A}$ 
(a list of $K$ items), and receives user feedback $r_t$.

\subsubsection{Latent Preference Manifold Hypothesis}

A critical challenge in real-world systems is that the observed user state 
$\tilde{s}_t$ is not a faithful representation of the user's true preference. 
User interactions are systematically corrupted by exposure bias, conformity effects, 
and popularity-driven feedback loops.

We hypothesize that true user preferences lie on a low-dimensional latent manifold 
$\mathcal{M} \subset \mathbb{R}^d$, while the observed interaction state is a corrupted 
projection off this manifold:
\begin{equation}
\tilde{s}_t = \mathcal{M}(s_t^*) + \zeta_{pop},
\end{equation}
where $s_t^* \in \mathcal{M}$ denotes the latent preference state and 
$\zeta_{pop}$ represents structured epistemic noise induced by popularity bias 
(Challenge \textbf{C1}).

This implies that $\tilde{s}_t$ does not preserve the intrinsic geometry of user 
preferences, and directly learning policies on $\tilde{s}_t$ leads to systematic 
overfitting to spurious popularity signals.

Our objective is therefore to recover a purified latent representation 
$\hat{s}_t \approx s_t^*$ that captures genuine preference structure while being 
invariant to exposure-driven distortions.

\subsection{DSRM: Diffusion-based State Purification}

To address Challenge \textbf{C2} (signal-noise entanglement), we formulate state 
purification as a conditional generative modeling problem. 
The Denoising State Representation Module (DSRM) learns a probabilistic 
projection operator:
\[
\Pi_\theta : \mathbb{R}^d \rightarrow \mathcal{M},
\]
which maps corrupted states back onto the latent preference manifold.

\subsubsection{Forward Diffusion Process}

We define a forward diffusion process that progressively injects Gaussian noise 
into the state embedding over $K$ steps. Given an initial encoded state 
$s^{(0)} = \tilde{s}_t$, the distribution at step $k$ is:
\begin{equation}
q(s^{(k)} | s^{(0)}) = 
\mathcal{N}(s^{(k)}; \sqrt{\bar{\alpha}_k}s^{(0)}, (1-\bar{\alpha}_k)\mathbf{I}),
\end{equation}
where $\bar{\alpha}_k$ follows a predefined variance schedule. 
This process simulates the progressive degradation of preference signals under 
extreme exposure bias.

\subsubsection{Reverse Diffusion: Manifold Projection}

The reverse diffusion learns to invert the corruption process by approximating:
\[
p_\theta(s^{(k-1)} | s^{(k)}, \tilde{s}_t),
\]
which iteratively reconstructs the clean preference structure:
\begin{equation}
s^{(k-1)} = \frac{1}{\sqrt{\alpha_k}}
\left( s^{(k)} -
\frac{1-\alpha_k}{\sqrt{1-\bar{\alpha}_k}}
\epsilon_\theta(s^{(k)}, k, \tilde{s}_t) \right)
+ \sigma_k \mathbf{z}.
\end{equation}

From a geometric perspective, this reverse diffusion implements an 
\textbf{iterative probabilistic manifold projection}, enabling recovery of 
low-entropy preference representations while preserving fine-grained behavioral 
signals.

After $K$ reverse steps, we obtain the purified state:
\[
\hat{s}_t = \Pi_\theta(\tilde{s}_t),
\]
which serves as the decision-relevant representation for policy learning.

\subsection{HRL: Hierarchical Constrained Control}

To resolve Challenge \textbf{C3} (temporal objective conflict), we decompose 
decision-making into a hierarchical control structure that decouples 
macro-level fairness regulation from micro-level engagement optimization.

\subsubsection{High-Level Manager: Fairness Regulation}

The high-level policy $\pi_h(z_t | \hat{s}_t)$ outputs a strategic control variable 
$z_t = [\omega_{acc}, \omega_{fair}] \in \mathbb{R}^2$, which dynamically defines 
the fairness constraint for the current step.

The manager optimizes an ecosystem-level objective:
\begin{equation}
r_t^h = r_t + \lambda \cdot \text{Fair}(\mathcal{L}_t),
\end{equation}
where $\text{Fair}(\cdot)$ measures long-term exposure equity (e.g., negative Gini 
coefficient of item exposure).

This induces a constrained optimization problem for the low-level policy:
\begin{equation}
\max_{\pi_l} \ \mathbb{E}[R_{acc}]
\quad \text{s.t.} \quad 
\mathbb{E}[R_{fair}] \geq z_t,
\end{equation}
where the constraint $z_t$ is dynamically adjusted by the manager.

\subsubsection{Low-Level Worker: Utility Optimization}

The low-level policy $\pi_l(a_t | \hat{s}_t, z_t)$ selects items within the feasible 
fairness manifold. For each candidate item $i$, it computes:
\begin{equation}
\Psi(i; \hat{s}_t, z_t) =
\omega_{acc} \cdot \text{Sim}(\hat{s}_t, \mathbf{e}_i)
- \omega_{fair} \cdot \log(\text{pop}_i),
\end{equation}
where $\text{Sim}(\cdot)$ denotes cosine similarity in purified space and 
$\text{pop}_i$ is the item's global popularity.

This formulation ensures that the worker maximizes immediate engagement 
while remaining strictly constrained by the manager’s fairness regulation.

\subsection{Joint Optimization Strategy}

We adopt a \textbf{Purify-then-Decouple} training paradigm.

\textbf{Stage I: DSRM Pre-training.}  
The diffusion module is trained via noise reconstruction loss:
\begin{equation}
\mathcal{L}_{DSRM} =
\mathbb{E}\|\epsilon - \epsilon_\theta(s^{(k)}, k, \tilde{s}_t)\|^2.
\end{equation}

\textbf{Stage II: Hierarchical Policy Learning.}  
With DSRM fixed, both hierarchical policies are optimized via PPO. 
Hierarchical gradient separation mitigates learning interference, allowing 
the fairness controller to converge stably under high-variance feedback.

\section{Experiments}

In this section, we conduct extensive experiments to systematically evaluate the proposed DSRM-HRL framework and answer the following research questions:
\begin{itemize}
    \item \textbf{RQ1}: How does DSRM-HRL perform compared with state-of-the-art reinforcement learning methods in interactive recommendation environments?
    \item \textbf{RQ2}: How do different components of DSRM-HRL contribute to its overall performance?
    \item \textbf{RQ3}: How sensitive is DSRM-HRL to the number of diffusion steps under different environments?
    \item \textbf{RQ4}: What is the computational overhead of DSRM-HRL compared with existing methods in terms of training efficiency?
    \item \textbf{RQ5}: Does the hierarchical decoupling and diffusion-based state purification improve training stability and convergence behavior?
\end{itemize}

\subsection{Experimental Setup}

We introduce the experimental setup, including the simulation environment, evaluation metrics, and baseline methods.

\subsubsection{Simulation Environment}

To evaluate interactive recommendation policies in a controllable and reproducible setting under the accuracy--fairness trade-off, we adopt the interactive simulator \textbf{KuaiSim}~\cite{b17}. 
Unlike static datasets, KuaiSim models dynamic user feedback and long-term satisfaction, enabling fine-grained analysis of sequential decision-making.

KuaiSim is constructed based on two real-world short video datasets:

\textbf{KuaiRec}~\cite{b18}: a fully observed user--item interaction matrix with 1,411 users and 3,327 videos, supporting accurate modeling of user preferences.

\textbf{KuaiRand-Pure}~\cite{b19}: a larger dataset with 27,285 users, 7,583 items, and over 1.18 million interactions collected under randomized exposure, suitable for evaluating generalization.

Following the official protocol, we split the datasets into independent training and testing environments, as summarized in Table~\ref{tab:dataset}. Separate simulators are constructed for each environment, and the test environments are strictly held out during training, ensuring that all recommendation models are evaluated under fully unseen conditions.

\begin{table}[t]
\centering
\caption{Statistics of the training and testing environments.}
\label{tab:dataset}
\begin{tabular}{l l r r r}
\hline
Dataset & Usage & Users & Items & Interactions \\
\hline
\multirow{2}{*}{KuaiRec} 
& Train & 7,176 & 10,728 & 12,530.8k \\
& Test  & 1,411 & 3,327  & 4,676.5k \\
\hline
\multirow{2}{*}{KuaiRand} 
& Train & 26,285 & 7,551 & 1,436.6k \\
& Test  & 27,285 & 7,583 & 1,186.1k \\
\hline
\end{tabular}
\end{table}

To study item-side fairness and interaction dynamics, we integrate two mechanisms:

\textbf{(1) Popularity-based grouping.}  
Following~\cite{b20}, items are divided into popular (top 20\%) and long-tail (remaining 80\%) groups based on exposure frequency. We set the maximum session length (\textit{Max Len}) to 30 and 50.

\textbf{(2) Fairness-aware abandonment.}  
Inspired by~\cite{b12, b22}, if the system repeatedly recommends popular items, user satisfaction decreases and the session terminates early. This enforces an explicit trade-off between short-term relevance and long-term fairness, providing a realistic testbed for DSRM-HRL.

\subsubsection{Evaluation Metrics}

We adopt four complementary metrics: \textbf{(1)Interaction Length (Len)}:  
Average number of interaction steps per session. Under fairness-aware abandonment, Len directly reflects long-term user satisfaction and strategic effectiveness. \textbf{(2)Cumulative Reward ($R_{cum}$)}:  Total reward accumulated over a session, measuring long-term system utility. \textbf{(3)Single-step Reward ($R_{each}$)}:  Average immediate reward, reflecting short-term recommendation relevance. High $R_{each}$ alone may indicate myopic exploitation. \textbf{(4)Absolute Difference (AD)}\cite{ad}:  Group fairness metric measuring exposure disparity:
\[
AD = \left| f(G_{pop}) - f(G_{longtail}) \right|,
\]
where $G$ denotes the average proportion of items in the group that appearing on recommendation lists. Lower AD indicates more balanced exposure.

\subsubsection{Baselines}

We compare DSRM-HRL with two categories of baselines.
\textbf{(1)General RL methods} (no fairness modeling):  
A2C~\cite{b23},  
TD3~\cite{b24},  
BCQ~\cite{b25}.
\textbf{(2)Fairness-aware RL methods}:  
MOFIR~\cite{b20},  
DORL~\cite{b27},  
DNAIR~\cite{b12},  
SAC4IR~\cite{b29}.
All baselines are carefully tuned using grid search.  
Results are averaged over three random seeds \{11, 15, 19\}.
Our code will be open-sourced upon publication.

\subsection{Overall Performance (RQ1)}
\begin{table*}[htbp]
\centering
\caption{Overall Experimental Results}
\label{tab:overall}
\resizebox{\textwidth}{!}{
\begin{tabular}{llcccccccc}
\toprule
\multirow{2}{*}{Dataset} & \multirow{2}{*}{Method} 
& \multicolumn{4}{c}{Max Len = 30} 
& \multicolumn{4}{c}{Max Len = 50} \\
\cmidrule(lr){3-6} \cmidrule(lr){7-10}
 &  & Len & $R_{each}$ & $R_{cum}$ & AD & Len & $R_{each}$ & $R_{cum}$ & AD \\
\midrule
\multirow{8}{*}{KuaiRec}
&A2C    & 13.433$\pm$0.236 & 0.620$\pm$0.056 & 8.339$\pm$0.903  & 0.498$\pm$0.085 & 22.633$\pm$0.386 & 0.680$\pm$0.057 & 15.422$\pm$1.727 & 0.577$\pm$0.137 \\
&TD3    & 15.567$\pm$1.733 & 0.801$\pm$0.029 & 12.007$\pm$1.260 & 0.428$\pm$0.073 & 32.267$\pm$6.712 & 0.850$\pm$0.022 & 26.627$\pm$6.444 & 0.293$\pm$0.081 \\
&BCQ    & 20.800$\pm$0.616 & 0.757$\pm$0.004 & 15.036$\pm$0.314 & 0.021$\pm$0.003 & 34.067$\pm$0.759 & 0.825$\pm$0.003 & 28.043$\pm$0.343 & 0.037$\pm$0.007 \\
&MOFIR  & 13.733$\pm$1.406 & 0.577$\pm$0.092 & 7.852$\pm$1.816  & 0.668$\pm$0.130 & 22.933$\pm$1.901 & 0.616$\pm$0.093 & 14.596$\pm$3.295 & 0.668$\pm$0.120 \\
&DORL   & 14.700$\pm$2.376 & 0.579$\pm$0.061 & 7.490$\pm$0.889  & 0.466$\pm$0.236 & 22.133$\pm$0.939 & 0.667$\pm$0.100 & 14.564$\pm$3.407 & 0.548$\pm$0.041 \\
&DNaIR    & 21.733$\pm$2.713 & 0.809$\pm$0.014 & 17.699$\pm$2.704 & 0.083$\pm$0.056 & 23.767$\pm$0.330 & 0.809$\pm$0.023 & 19.325$\pm$0.440 & 0.660$\pm$0.026 \\
&SAC4IF & 21.967$\pm$0.450 & 0.828$\pm$0.026 & 18.116$\pm$1.236 & 0.013$\pm$0.003 & 33.633$\pm$5.329 & 0.865$\pm$0.029 & 29.470$\pm$5.577 & 0.151$\pm$0.196 \\
\cmidrule(lr){2-10}
&Ours   & \textbf{26.600$\pm$0.816} & \textbf{0.917$\pm$0.010} & \textbf{23.752$\pm$1.043} & \textbf{0.008$\pm$0.003} & \textbf{34.850$\pm$0.350} & \textbf{0.870$\pm$0.001} & \textbf{30.419$\pm$0.485} & 0.044$\pm$0.011 \\
\midrule
\multirow{8}{*}{KuaiRand-Pure}
&A2C    & 13.667$\pm$0.125 & 0.730$\pm$0.014 & 9.813$\pm$0.300  & 0.923$\pm$0.013 & 23.167$\pm$0.125 & 0.819$\pm$0.008 & 18.271$\pm$0.671 & 0.925$\pm$0.011 \\
&TD3    & 16.233$\pm$2.311 & 0.810$\pm$0.004 & 12.468$\pm$1.913 & 0.344$\pm$0.175 & 31.233$\pm$7.811 & 0.842$\pm$0.019 & 25.811$\pm$7.307 & 0.298$\pm$0.208 \\
&BCQ    & 17.133$\pm$0.330 & 0.789$\pm$0.004 & 12.925$\pm$0.117 & 0.295$\pm$0.035 & 25.900$\pm$2.974 & 0.838$\pm$0.008 & 20.395$\pm$2.474 & 0.514$\pm$0.215 \\
&MOFIR  & 13.900$\pm$0.216 & 0.745$\pm$0.015 & 10.172$\pm$0.333 & 0.813$\pm$0.077 & 23.700$\pm$0.920 & 0.833$\pm$0.009 & 18.486$\pm$0.487 & 0.805$\pm$0.088 \\
&DORL   & 13.700$\pm$0.082 & 0.727$\pm$0.017 & 9.724$\pm$0.236  & 0.946$\pm$0.013 & 23.100$\pm$0.082 & 0.812$\pm$0.003 & 17.766$\pm$0.236 & 0.954$\pm$0.015 \\
&DNaIR    & 16.267$\pm$2.176 & 0.815$\pm$0.016 & 12.832$\pm$1.767 & 0.559$\pm$0.349 & 31.800$\pm$7.105 & 0.847$\pm$0.013 & 26.299$\pm$6.319 & 0.443$\pm$0.286 \\
&SAC4IF & 14.000$\pm$0.000 & 0.788$\pm$0.001 & 10.862$\pm$0.069 & 0.674$\pm$0.009 & 28.433$\pm$6.557 & 0.834$\pm$0.006 & 23.123$\pm$5.776 & 0.394$\pm$0.265 \\
\cmidrule(lr){2-10}
&Ours   & \textbf{24.067$\pm$1.396} & \textbf{0.849$\pm$0.007} & \textbf{19.750$\pm$1.392} & \textbf{0.011$\pm$0.003} & \textbf{35.133$\pm$0.858} & \textbf{0.848$\pm$0.001} & \textbf{29.086$\pm$0.505} & \textbf{0.024$\pm$0.009} \\
\bottomrule
\end{tabular}
}
\end{table*}
Table~\ref{tab:overall} reports the overall performance. DSRM-HRL consistently achieves the best performance across most metrics on both datasets, demonstrating a superior accuracy--fairness trade-off. For example, on KuaiRec with Max Len = 30, DSRM-HRL achieves an average interaction length of 26.600, improving over the strongest fairness baseline SAC4IR (21.967) by 21.1\%, and over the best general RL baseline BCQ (20.800) by 27.9\%. These gains indicate that DSRM-HRL effectively avoids fairness-triggered abandonment and maximizes long-term retention. Importantly, this improvement is not achieved at the cost of short-term relevance.DSRM-HRL also achieves the highest $R_{each}$ and \textbf{competitive AD}, maintaining a superior harmony between accuracy and fairness. While BCQ achieves a low AD due to its conservative policy constraints, it fails to match the high long-term engagement (Len) of our approach. In contrast, standard RL methods (e.g., A2C, TD3) suffer from severe exposure imbalance (high AD), leading to early termination.These results answer \textbf{RQ1}: DSRM-HRL significantly outperforms both general and fairness-aware RL baselines in interactive environments.

\subsection{Ablation Study (RQ2)}
\begin{table*}[htbp]
\centering
\caption{Ablation Study}
\label{tab:ablation}
\resizebox{\textwidth}{!}{
\begin{tabular}{llcccccccc}
\toprule
\multirow{2}{*}{Dataset} & \multirow{2}{*}{Method} 
& \multicolumn{4}{c}{Max Len = 30} 
& \multicolumn{4}{c}{Max Len = 50} \\
\cmidrule(lr){3-6} \cmidrule(lr){7-10}
 &  & Len & $R_{each}$ & $R_{cum}$ & AD & Len & $R_{each}$ & $R_{cum}$ & AD \\
\midrule
\multirow{6}{*}{KuaiRec}
&FLAT   & 24.333$\pm$0.741 & 0.911$\pm$0.003 & 22.081$\pm$1.008 & 0.013$\pm$0.005 & 31.267$\pm$4.658 & 0.870$\pm$0.045 & 26.691$\pm$2.693 & 0.222$\pm$0.291 \\
&HRL    & 23.167$\pm$0.411 & 0.895$\pm$0.016 & 19.942$\pm$0.723 & 0.013$\pm$0.001 & 33.400$\pm$0.566 & 0.863$\pm$0.005 & 28.055$\pm$0.396 & 0.016$\pm$0.002 \\
&+RCE   & 15.167$\pm$0.386 & 0.653$\pm$0.004 & 9.588$\pm$0.184  & 0.032$\pm$0.001 & 25.500$\pm$0.852 & 0.737$\pm$0.002 & 17.528$\pm$0.104 & 0.029$\pm$0.004 \\
&+TCE   & 15.867$\pm$0.403 & 0.746$\pm$0.007 & 11.491$\pm$0.353 & 0.030$\pm$0.002 & 25.800$\pm$0.497 & 0.793$\pm$0.012 & 20.293$\pm$0.444 & 0.040$\pm$0.005 \\
&+BOD   & 15.867$\pm$1.793 & 0.761$\pm$0.033 & 11.740$\pm$2.156 & 0.479$\pm$0.303 & 28.800$\pm$3.910 & 0.827$\pm$0.049 & 23.114$\pm$3.816 & 0.209$\pm$0.252 \\
\cmidrule(lr){2-10}
&+DSRM  & \textbf{26.600$\pm$0.816} & \textbf{0.917$\pm$0.010} & \textbf{23.752$\pm$1.043} & \textbf{0.008$\pm$0.003} & \textbf{34.850$\pm$0.350} & \textbf{0.870$\pm$0.001} & \textbf{30.419$\pm$0.485} & 0.044$\pm$0.011 \\
\midrule
\multirow{6}{*}{KuaiRand-Pure}
&FLAT   & 21.767$\pm$2.816 & 0.830$\pm$0.005 & 18.105$\pm$2.531 & 0.178$\pm$0.128 & 30.733$\pm$2.444 & 0.813$\pm$0.003 & 24.738$\pm$2.083 & 0.136$\pm$0.127 \\
&HRL    & 19.200$\pm$3.608 & 0.841$\pm$0.008 & 15.678$\pm$2.900 & 0.311$\pm$0.420 & 27.533$\pm$4.997 & 0.846$\pm$0.010 & 22.460$\pm$4.351 & 0.593$\pm$0.409 \\
&+RCE   & 15.100$\pm$0.082 & 0.765$\pm$0.002 & 11.267$\pm$0.020 & 0.047$\pm$0.006 & 25.833$\pm$0.377 & 0.835$\pm$0.001 & 20.428$\pm$0.208 & 0.035$\pm$0.005 \\
&+TCE   & 15.333$\pm$0.262 & 0.785$\pm$0.005 & 11.448$\pm$0.074 & 0.041$\pm$0.004 & 25.600$\pm$0.216 & 0.833$\pm$0.003 & 20.217$\pm$0.165 & 0.034$\pm$0.005 \\
&+BOD   & 14.233$\pm$0.170 & 0.780$\pm$0.011 & 10.865$\pm$0.179 & 0.747$\pm$0.055 & 25.533$\pm$1.112 & 0.838$\pm$0.005 & 20.316$\pm$1.355 & 0.577$\pm$0.167 \\
\cmidrule(lr){2-10}
&+DSRM  & \textbf{24.067$\pm$1.396} & \textbf{0.849$\pm$0.007} & \textbf{19.750$\pm$1.392} & \textbf{0.011$\pm$0.003} & \textbf{35.133$\pm$0.858} & \textbf{0.848$\pm$0.001} & \textbf{29.086$\pm$0.505} & \textbf{0.024$\pm$0.009} \\
\bottomrule
\end{tabular}
}
\end{table*}
To analyze the contribution of individual components, we evaluate the following variants:
(1)\textbf{FLAT}: DSRM with a single-level RL policy;
(2)\textbf{HRL}: Hierarchical RL without denoising;
(3)\textbf{HRL + RCE/TCE/BOD}: HRL combined with heuristic denoising methods~\cite{b30,b31}.

Results in Table~\ref{tab:ablation} reveal strong synergistic effects. First, FLAT consistently underperforms DSRM-HRL. For example, on KuaiRec (Max Len = 30), FLAT achieves Len = 24.333, while DSRM-HRL reaches 26.600. This indicates that even with purified states, single-policy RL struggles to reconcile conflicting objectives. Second, HRL without DSRM performs worse than the full model, confirming that hierarchical control alone is insufficient when states are corrupted. Third, combining HRL with heuristic denoising methods leads to severe degradation. For instance, HRL+RCE on KuaiRec yields Len = 15.167, far below HRL (23.167). This suggests that traditional denoising methods rely on rigid assumptions that may suppress valuable exploration signals and distort state representations in dynamic RL settings. In contrast, DSRM learns noise patterns from data and performs iterative probabilistic denoising, preserving decision-relevant information while removing spurious popularity bias. These findings answer \textbf{RQ2}: both diffusion-based state purification and hierarchical decoupling are necessary.

\subsection{Diffusion Step Sensitivity (RQ3)}

\begin{figure}[!t]
\centering

\begin{minipage}[t]{.48\linewidth}
  \centering
  \includegraphics[width=\linewidth]{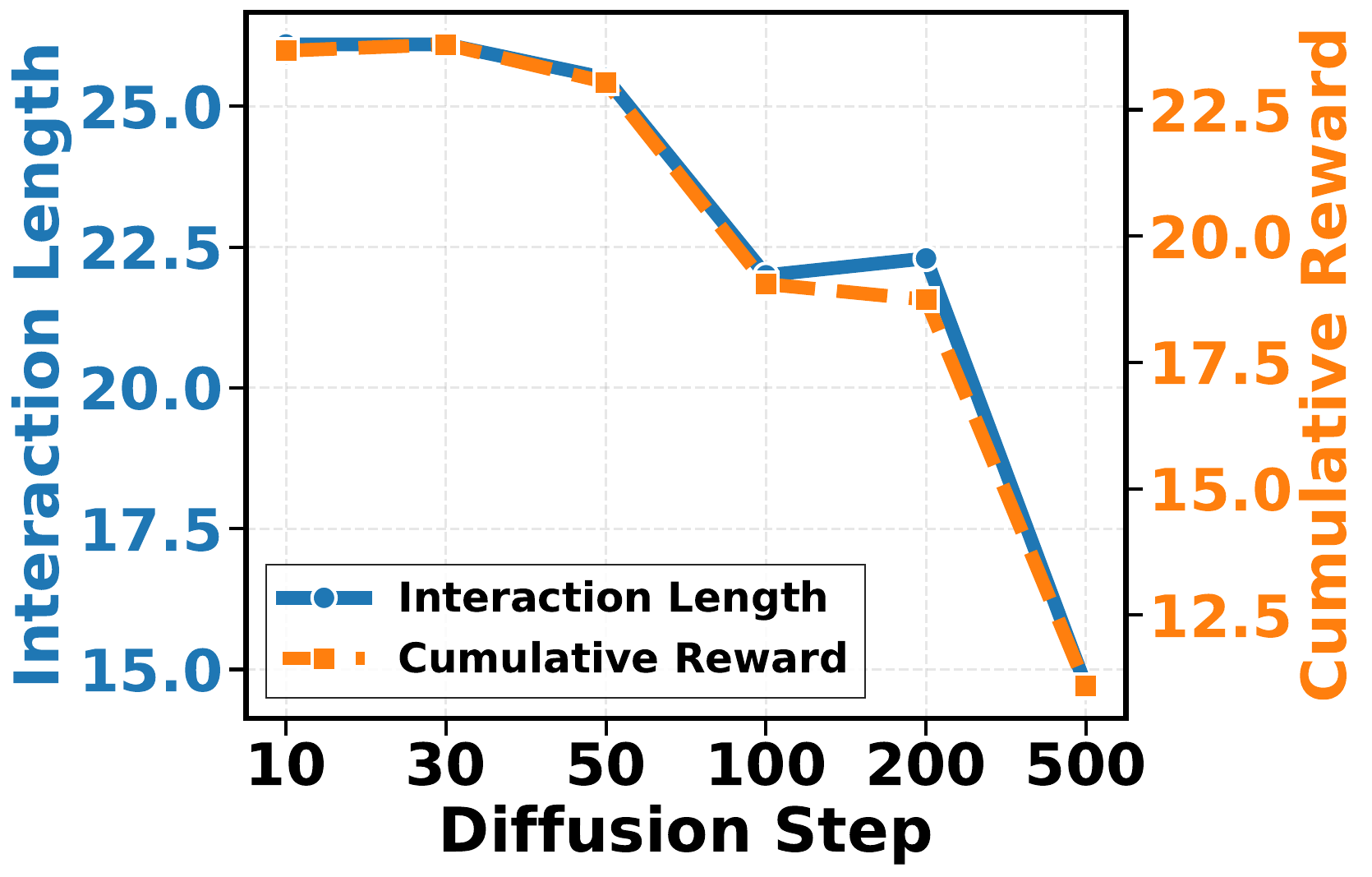}
\end{minipage}
\begin{minipage}[t]{.48\linewidth}
  \centering
  \includegraphics[width=\linewidth]{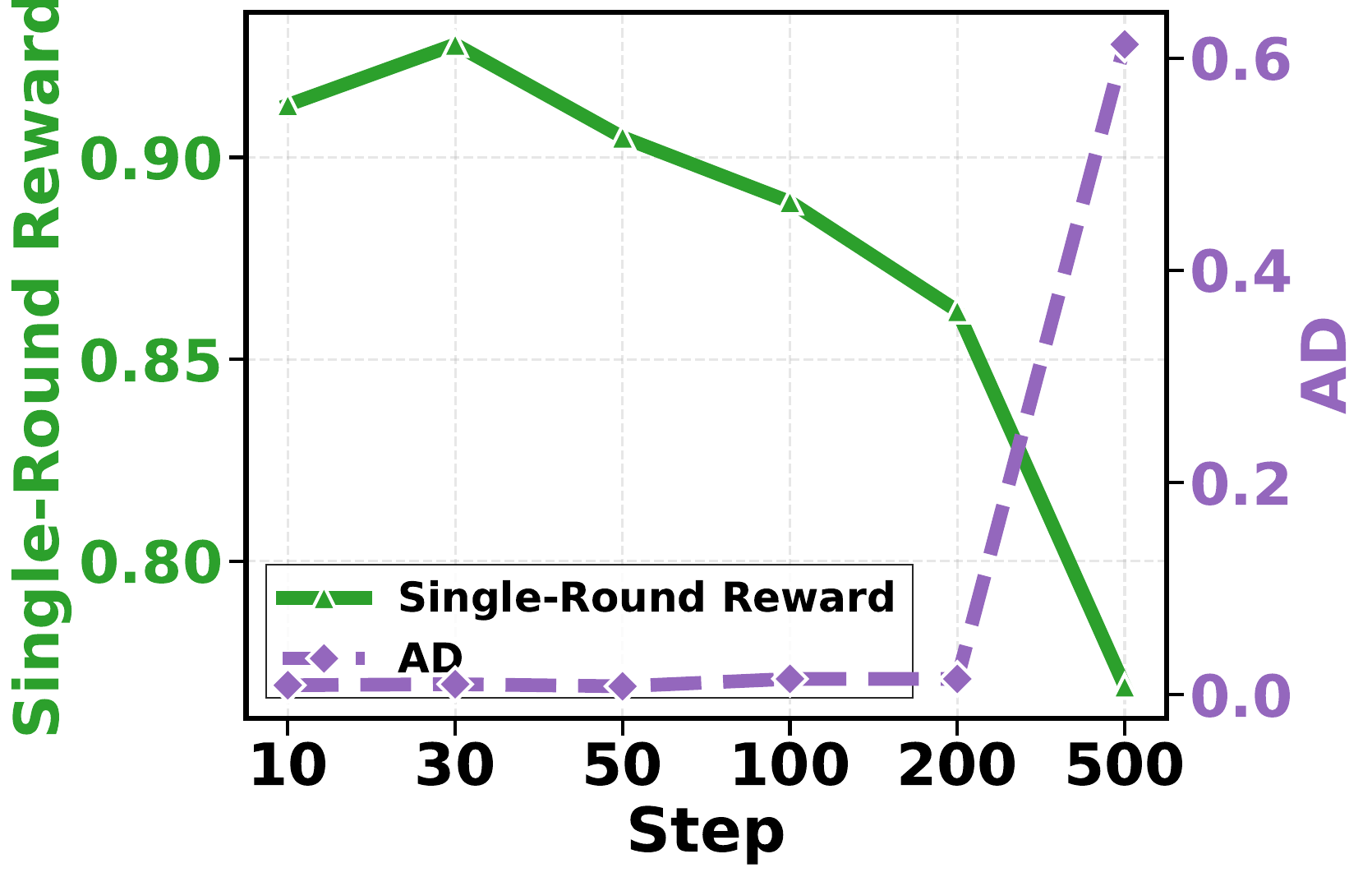}
\end{minipage}
\\[4pt]
{\small (a) Kuairec}

\vspace{8pt}

\begin{minipage}[t]{.48\linewidth}
  \centering
  \includegraphics[width=\linewidth]{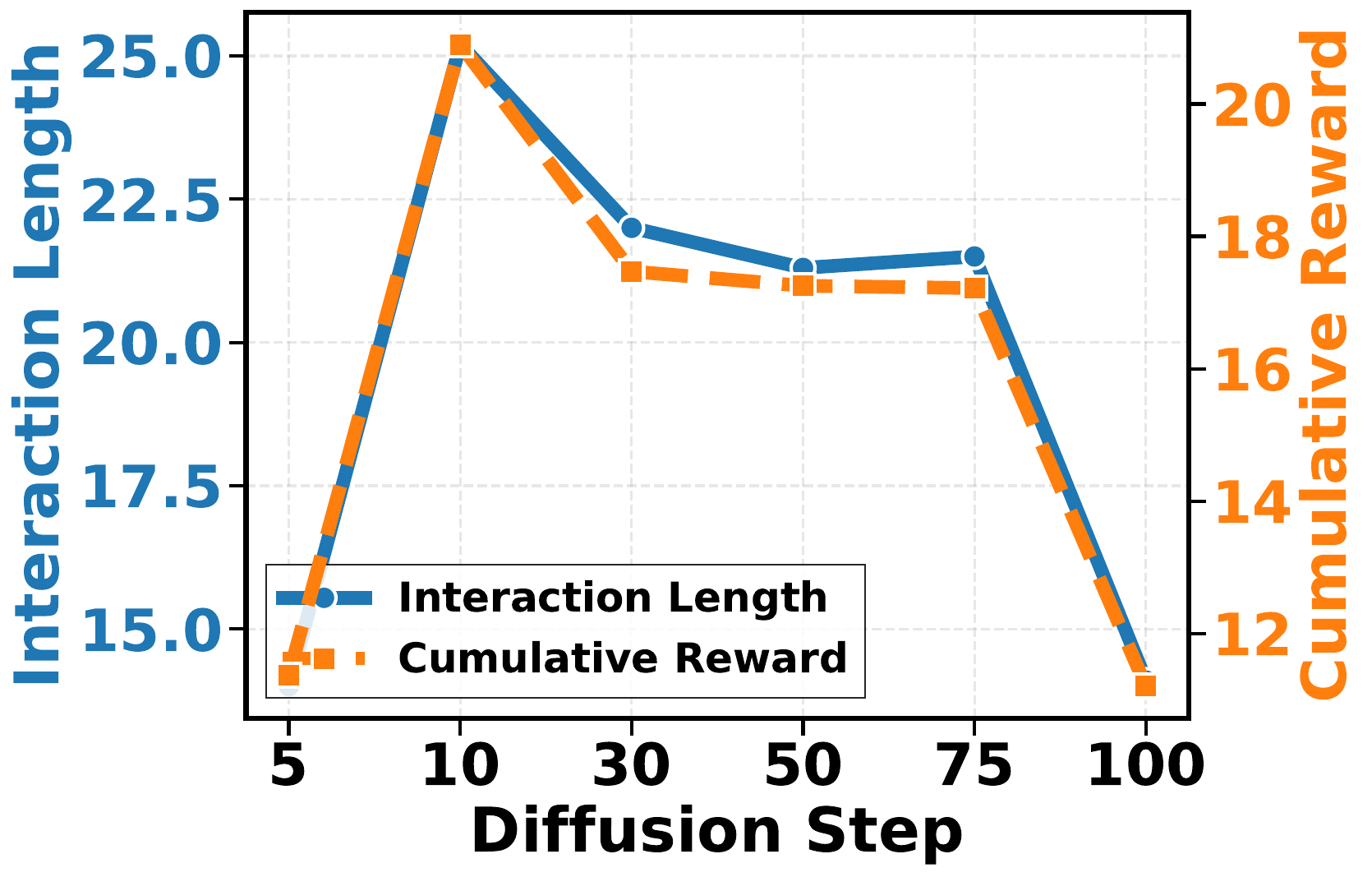}
\end{minipage}
\begin{minipage}[t]{.48\linewidth}
  \centering
  \includegraphics[width=\linewidth]{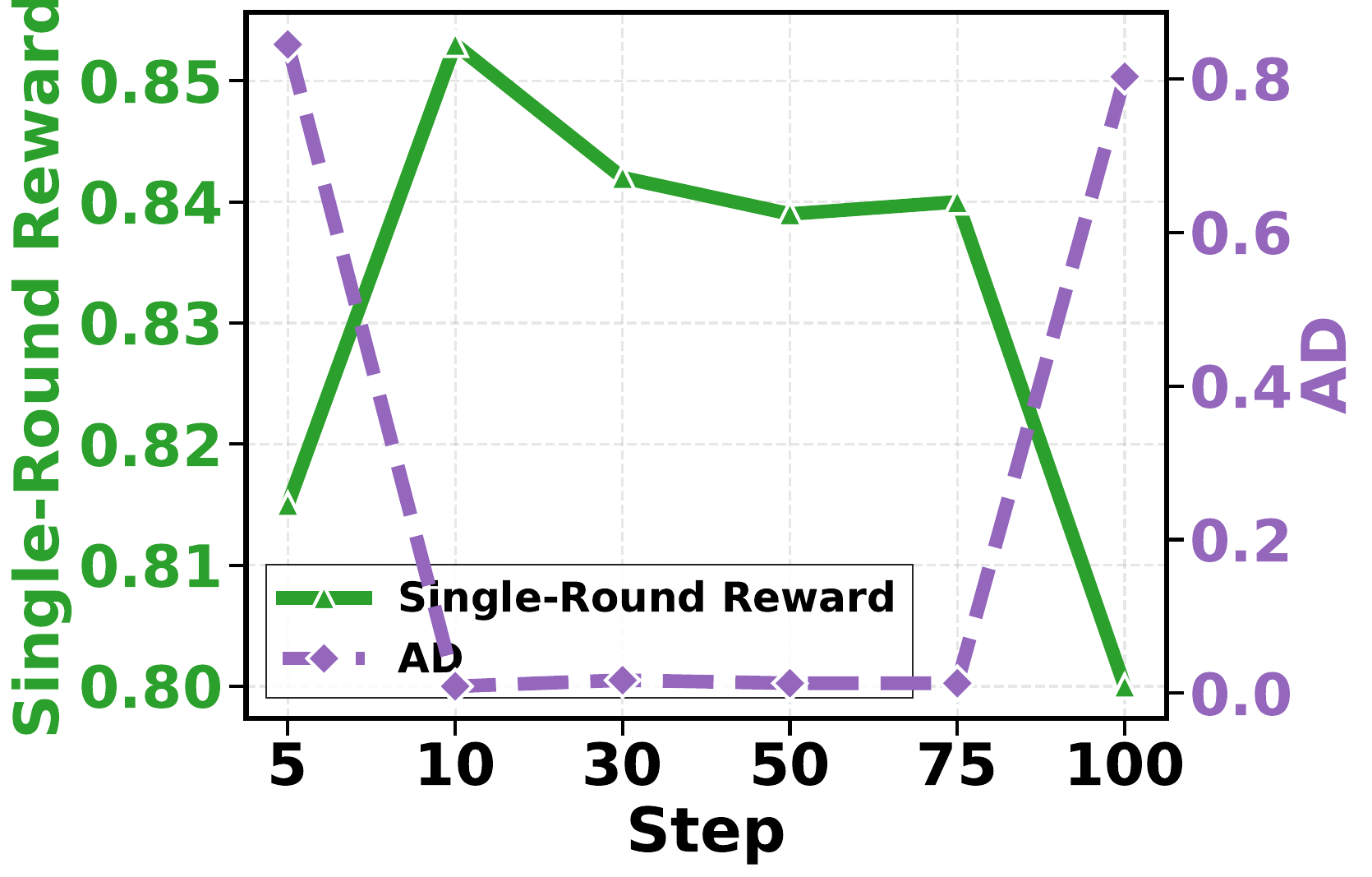}
\end{minipage}
\\[4pt]
{\small (b) KuaiRand}

\caption{Comparison of different diffusion step.}
\label{fig:diffusion}
\end{figure}

We vary the number of diffusion steps from 5 to 500. As shown in Figure~\ref{fig:diffusion}, performance peaks at moderate values (10--30) and deteriorates for large steps. On KuaiRec, increasing steps from 10 to 30 slightly improves Len and $R_{cum}$, while AD remains low. However, when steps increase to 100 or 500, Len drops sharply (e.g., to 14.7 at 500). This reflects an over-smoothing effect: excessive denoising removes not only noise but also personalized preference signals, resulting in low-entropy but information-poor representations. Similar but more pronounced trends are observed on KuaiRand-Pure, where sparse data requires stronger but carefully controlled denoising. These results answer \textbf{RQ3}: DSRM requires a balanced number of diffusion steps to trade off noise removal and information preservation.

\subsection{Computational Efficiency (RQ4)}

\begin{table}[t]
\centering
\caption{Training time comparison on KuaiRec (Max Len=30, Episode=20000).}
\label{tab:time}
\begin{tabular}{l c}
\toprule
Method & Time (seconds) \\
\midrule
A2C    & 7512.144 \\
SAC4IF    & 6754.545 \\
DNaIR  & 5934.623 \\
HRL   & 8462.525 \\
RCE    & 29919.296 \\
\cmidrule{1-2}
DSRM-HRL (Ours) & 15909.175 \\
\bottomrule
\end{tabular}
\end{table}

Table~\ref{tab:time} compares the training time of some methods on KuaiRec under the same budget (Max Len = 30, Episode = 20,000). 
As expected, DSRM-HRL introduces additional computational overhead due to diffusion-based state purification and hierarchical control, requiring 15,909 seconds, which is about 2.1$\times$ that of DNaIR and 2.3$\times$ that of SAC4IF. However, this cost is substantially lower than heuristic denoising baselines such as RCE (29,919s), indicating that the learned diffusion process is significantly more efficient than handcrafted denoising strategies. More importantly, the moderate overhead yields substantial gains in both long-term utility and fairness (Table~\ref{tab:overall}), demonstrating a favorable efficiency--performance trade-off. 
Since diffusion operates on compact latent states rather than raw features, the inference overhead remains practical for real-world deployment. These results answer \textbf{RQ4}: DSRM-HRL achieves strong performance gains with acceptable and well-controlled computational cost.

\subsection{Convergence and Training Stability (RQ5)}
\begin{figure}[!t]
    \centering
    \begin{subfigure}{0.96\linewidth}
        \includegraphics[width=\textwidth]{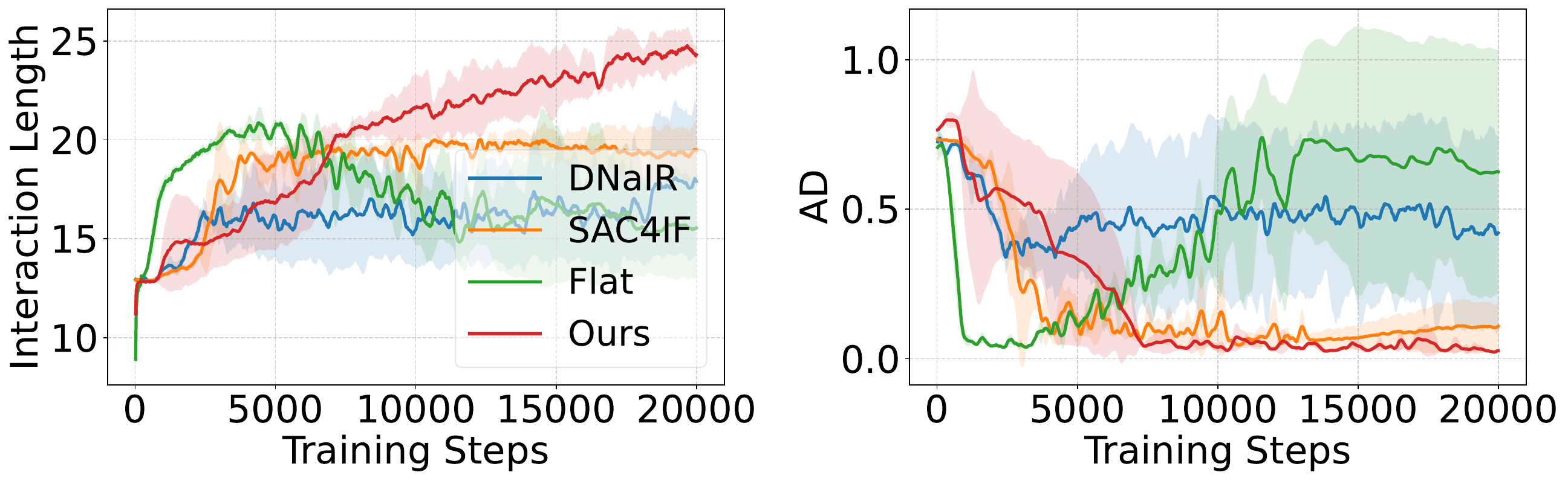}
        \caption{KuaiRec}
        \label{fig:training_rec}
    \end{subfigure}
    
    \begin{subfigure}{0.96\linewidth}
        \includegraphics[width=\textwidth]{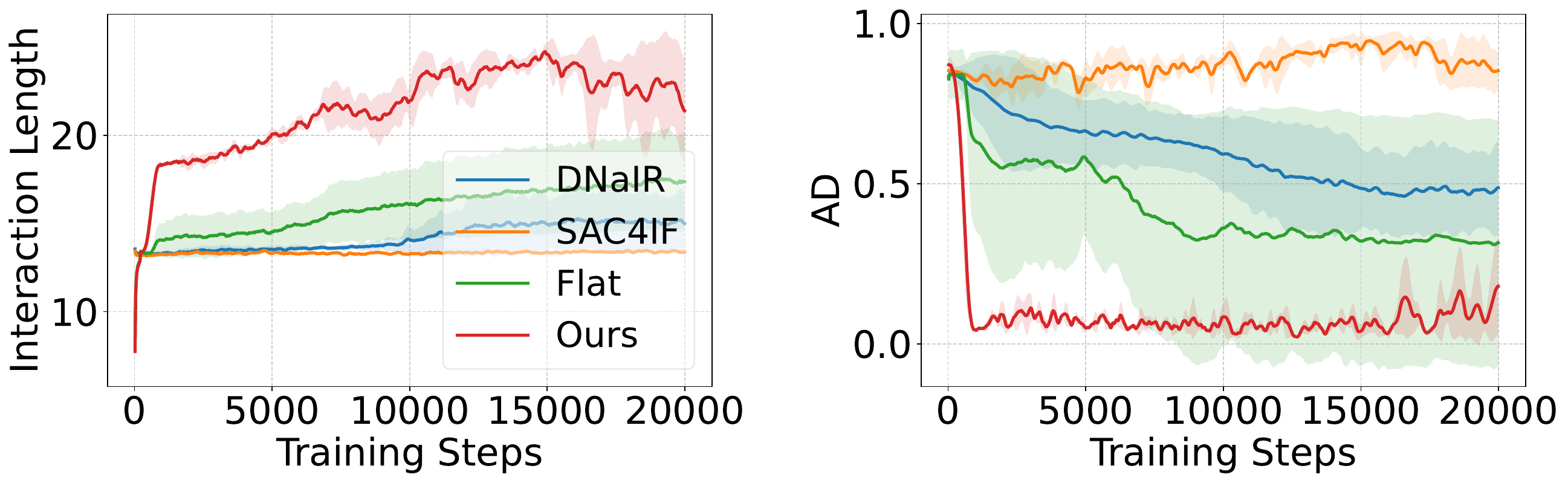}
        \caption{KuaiRand}
        \label{fig:training_rand}
    \end{subfigure}
    \caption{Training dynamics on KuaiRec and KuaiRand-Pure (Max Len = 30). 
    We report learning curves of Interaction Length (left) and AD (right) over 20,000 training steps averaged across three random seeds. 
    DSRM-HRL exhibits smoother convergence and lower variance than baselines, indicating improved training stability in non-stationary environments.}

\end{figure}

Figures~\ref{fig:training_rec} and~\ref{fig:training_rand} illustrate the training dynamics of Interaction Length and AD on KuaiRec and KuaiRand-Pure. Across both datasets, DSRM-HRL demonstrates consistently smoother convergence and significantly lower variance than all baselines. While flat RL methods and fairness-aware agents exhibit strong oscillations and frequent performance collapses, DSRM-HRL improves steadily and reaches stable high-performance regimes. In particular, Interaction Length under DSRM-HRL increases monotonically and saturates earlier, whereas AD rapidly decreases and remains consistently low. 
These results indicate that fairness constraints are internalized through hierarchical control rather than enforced by unstable external penalties. 
Overall, hierarchical decoupling together with diffusion-based state purification transforms the originally biased and non-stationary learning process into a more stable and learnable optimization problem. These findings answer \textbf{RQ5}: DSRM-HRL significantly improves convergence reliability and training stability in interactive recommendation.

\section{Conclusion}
In this work, we challenge the conventional wisdom that the accuracy--fairness conflict is an inherent policy trade-off. We argue it is primarily a \textbf{problem of noisy state estimation}. By introducing \textbf{DSRM-HRL}, we demonstrate that \textbf{fairness begins with state purification}. Our diffusion-based reconstruction recovers the latent preference manifold, while our hierarchical control decouples conflicting temporal objectives. Extensive experiments prove that this "purify-then-decouple" paradigm not only promotes equity but also drives long-term user retention, offering a robust path for responsible AI in sequential decision-making.

\bibliographystyle{ACM-Reference-Format}
\bibliography{ref}

@article{b1,
  title={Deep reinforcement learning for list-wise recommendations},
  author={Zhao, Xiangyu and Zhang, Liang and Xia, Long and Ding, Zhuoye and Yin, Dawei and Tang, Jiliang},
  journal={arXiv preprint arXiv:1801.00209},
  year={2017}
}

@inproceedings{b2,
  title={Stabilizing reinforcement learning in dynamic environment with application to online recommendation},
  author={Chen, Shi-Yong and Yu, Yang and Da, Qing and Tan, Jun and Huang, Hai-Kuan and Tang, Hai-Hong},
  booktitle={Proceedings of the 24th ACM SIGKDD International Conference on Knowledge Discovery \& Data Mining},
  pages={1187--1196},
  year={2018}
}

@book{b3,
  title={Recommender systems: Algorithms and applications},
  author={Kumar, P Pavan and Vairachilai, S and Potluri, Sirisha and Mohanty, Sachi Nandan},
  year={2021},
  publisher={CRC press}
}

@inproceedings{b4,
  title={Degenerate feedback loops in recommender systems},
  author={Jiang, Ray and Chiappa, Silvia and Lattimore, Tor and Gy{\"o}rgy, Andr{\'a}s and Kohli, Pushmeet},
  booktitle={Proceedings of the 2019 AAAI/ACM Conference on AI, Ethics, and Society},
  pages={383--390},
  year={2019}
}

@article{b5,
  title={Learning from noisy labels with deep neural networks: A survey},
  author={Song, Hwanjun and Kim, Minseok and Park, Dongmin and Shin, Yooju and Lee, Jae-Gil},
  journal={IEEE transactions on neural networks and learning systems},
  volume={34},
  number={11},
  pages={8135--8153},
  year={2022},
  publisher={IEEE}
}

@article{b6,
  title={Bias and debias in recommender system: A survey and future directions},
  author={Chen, Jiawei and Dong, Hande and Wang, Xiang and Feng, Fuli and Wang, Meng and He, Xiangnan},
  journal={ACM Transactions on Information Systems},
  volume={41},
  number={3},
  pages={1--39},
  year={2023},
  publisher={ACM New York, NY}
}

@inproceedings{b7,
  title={Controlling popularity bias in learning-to-rank recommendation},
  author={Abdollahpouri, Himan and Burke, Robin and Mobasher, Bamshad},
  booktitle={Proceedings of the eleventh ACM conference on recommender systems},
  pages={42--46},
  year={2017}
}

@inproceedings{b8,
  title={Item popularity and recommendation accuracy},
  author={Steck, Harald},
  booktitle={Proceedings of the fifth ACM conference on Recommender systems},
  pages={125--132},
  year={2011}
}

@inproceedings{b9,
  title={Towards long-term fairness in recommendation},
  author={Ge, Yingqiang and Liu, Shuchang and Gao, Ruoyuan and Xian, Yikun and Li, Yunqi and Zhao, Xiangyu and Pei, Changhua and Sun, Fei and Ge, Junfeng and Ou, Wenwu and others},
  booktitle={Proceedings of the 14th ACM international conference on web search and data mining},
  pages={445--453},
  year={2021}
}

@article{b10,
  title={Relieving popularity bias in interactive recommendation: A diversity-novelty-aware reinforcement learning approach},
  author={Shi, Xiaoyu and Liu, Quanliang and Xie, Hong and Wu, Di and Peng, Bo and Shang, MingSheng and Lian, Defu},
  journal={ACM Transactions on Information Systems},
  volume={42},
  number={2},
  pages={1--30},
  year={2023},
  publisher={ACM New York, NY, USA}
}

@article{b11,
  title={Cirec: Causal intervention-inspired policy learning to mitigate exposure bias for interactive recommendation},
  author={Zheng, Yongsen and Wang, Guohua and Qin, Jinghui and Chen, Ziliang and Lin, Junfan and Wei, Pengxu and Lin, Liang and Lam, Kwok-Yan},
  journal={IEEE Transactions on Knowledge and Data Engineering},
  year={2025},
  publisher={IEEE}
}

@article{b12,
  title={Maximum entropy policy for long-term fairness in interactive recommender systems},
  author={Shi, Xiaoyu and Liu, Quanliang and Xie, Hong and Bai, Yanan and Shang, Mingsheng},
  journal={IEEE Transactions on Services Computing},
  volume={17},
  number={3},
  pages={1029--1043},
  year={2024},
  publisher={IEEE}
}

@inproceedings{b13,
  title={Collaborative denoising auto-encoders for top-n recommender systems},
  author={Wu, Yao and DuBois, Christopher and Zheng, Alice X and Ester, Martin},
  booktitle={Proceedings of the ninth ACM international conference on web search and data mining},
  pages={153--162},
  year={2016}
}

@inproceedings{b14,
  title={Learning to reweight examples for robust deep learning},
  author={Ren, Mengye and Zeng, Wenyuan and Yang, Bin and Urtasun, Raquel},
  booktitle={International conference on machine learning},
  pages={4334--4343},
  year={2018},
  organization={PMLR}
}

@article{b15,
  title={Diffusion Models in Recommendation Systems: A Survey},
  author={Wei, Ting-Ruen and Fang, Yi},
  journal={arXiv preprint arXiv:2501.10548},
  year={2025}
}

@article{b16,
  title={Diffurec: A diffusion model for sequential recommendation},
  author={Li, Zihao and Sun, Aixin and Li, Chenliang},
  journal={ACM Transactions on Information Systems},
  volume={42},
  number={3},
  pages={1--28},
  year={2023},
  publisher={ACM New York, NY}
}

@article{b17,
  title={KuaiSim: A comprehensive simulator for recommender systems},
  author={Zhao, Kesen and Liu, Shuchang and Cai, Qingpeng and Zhao, Xiangyu and Liu, Ziru and Zheng, Dong and Jiang, Peng and Gai, Kun},
  journal={Advances in Neural Information Processing Systems},
  volume={36},
  pages={44880--44897},
  year={2023}
}

@article{b18,
  title={KuaiRec: A fully-observed dataset for recommender systems},
  author={Gao, Chongming and Li, Shijun and Lei, Wenqiang and Li, Biao and Jiang, Peng and Chen, Jiawei and He, Xiangnan and Mao, Jiaxin and Chua, Tat-Seng},
  journal={arXiv preprint arXiv:2202.10842},
  year={2022}
}

@inproceedings{b19,
  title={Kuairand: An unbiased sequential recommendation dataset with randomly exposed videos},
  author={Gao, Chongming and Li, Shijun and Zhang, Yuan and Chen, Jiawei and Li, Biao and Lei, Wenqiang and Jiang, Peng and He, Xiangnan},
  booktitle={Proceedings of the 31st ACM international conference on information \& knowledge management},
  pages={3953--3957},
  year={2022}
}

@inproceedings{b20,
  title={Toward pareto efficient fairness-utility trade-off in recommendation through reinforcement learning},
  author={Ge, Yingqiang and Zhao, Xiaoting and Yu, Lucia and Paul, Saurabh and Hu, Diane and Hsieh, Chu-Cheng and Zhang, Yongfeng},
  booktitle={Proceedings of the fifteenth ACM international conference on web search and data mining},
  pages={316--324},
  year={2022}
}

@inproceedings{b22,
  title={EasyRL4Rec: An Easy-to-use Library for Reinforcement Learning Based Recommender Systems},
  author={Yu, Yuanqing and Gao, Chongming and Chen, Jiawei and Tang, Heng and Sun, Yuefeng and Chen, Qian and Ma, Weizhi and Zhang, Min},
  booktitle={Proceedings of the 47th International ACM SIGIR Conference on Research and Development in Information Retrieval},
  pages={977--987},
  year={2024}
}

@inproceedings{b23,
  title={Asynchronous methods for deep reinforcement learning},
  author={Mnih, Volodymyr and Badia, Adria Puigdomenech and Mirza, Mehdi and Graves, Alex and Lillicrap, Timothy and Harley, Tim and Silver, David and Kavukcuoglu, Koray},
  booktitle={International conference on machine learning},
  pages={1928--1937},
  year={2016},
  organization={PmLR}
}

@inproceedings{b24,
  title={Addressing function approximation error in actor-critic methods},
  author={Fujimoto, Scott and Hoof, Herke and Meger, David},
  booktitle={International conference on machine learning},
  pages={1587--1596},
  year={2018},
  organization={PMLR}
}

@inproceedings{b25,
  title={Off-policy deep reinforcement learning without exploration},
  author={Fujimoto, Scott and Meger, David and Precup, Doina},
  booktitle={International conference on machine learning},
  pages={2052--2062},
  year={2019},
  organization={PMLR}
}

@inproceedings{b27,
  title={Alleviating matthew effect of offline reinforcement learning in interactive recommendation},
  author={Gao, Chongming and Huang, Kexin and Chen, Jiawei and Zhang, Yuan and Li, Biao and Jiang, Peng and Wang, Shiqi and Zhang, Zhong and He, Xiangnan},
  booktitle={Proceedings of the 46th international ACM SIGIR conference on research and development in information retrieval},
  pages={238--248},
  year={2023}
}

@article{b29,
  title={Relieving popularity bias in interactive recommendation: A diversity-novelty-aware reinforcement learning approach},
  author={Shi, Xiaoyu and Liu, Quanliang and Xie, Hong and Wu, Di and Peng, Bo and Shang, MingSheng and Lian, Defu},
  journal={ACM Transactions on Information Systems},
  volume={42},
  number={2},
  pages={1--30},
  year={2023},
  publisher={ACM New York, NY, USA}
}

@inproceedings{b30,
  title={Denoising implicit feedback for recommendation},
  author={Wang, Wenjie and Feng, Fuli and He, Xiangnan and Nie, Liqiang and Chua, Tat-Seng},
  booktitle={Proceedings of the 14th ACM international conference on web search and data mining},
  pages={373--381},
  year={2021}
}

@inproceedings{b31,
  title={Efficient bi-level optimization for recommendation denoising},
  author={Wang, Zongwei and Gao, Min and Li, Wentao and Yu, Junliang and Guo, Linxin and Yin, Hongzhi},
  booktitle={Proceedings of the 29th ACM SIGKDD conference on knowledge discovery and data mining},
  pages={2502--2511},
  year={2023}
}

@article{ad,
  title={A survey on the fairness of recommender systems},
  author={Wang, Yifan and Ma, Weizhi and Zhang, Min and Liu, Yiqun and Ma, Shaoping},
  journal={ACM Transactions on Information Systems},
  volume={41},
  number={3},
  pages={1--43},
  year={2023},
  publisher={ACM New York, NY}
}









\end{document}